# Relative velocity-based reward functions for crowd navigation of robots

Xiaoqing Yang, Fei Li

*Abstract*—The four-wheeled Mecanum robot is widely used in various industries due to its maneuverability and strong load capacity, which is suitable for performing precise transportation tasks in a narrow environment, but while the Mecanum wheel robot has mobility, it also consumes more energy than ordinary robots. The power consumed by the Mecanum wheel mobile robot varies enormously depending on their operating regimes and environments. Therefore, only knowing the working environment of the robot and the accurate power consumption model can we accurately predict the power consumption of the robot. In order to increase the appli-cable scenarios of energy consumption modeling for Mecanum wheel robots and improve the accuracy of energy consumption modeling, this paper focuses on various factors that affect the energy consumption of the Mecanum wheel robot, such as motor temperature, terrain, the center of gravity position, etc. The model is derived from the kinematic and kinetic model combined with electrical engineering and energy flow principles. The model has been simulated in MATLAB and experimentally validated with the four-wheeled Mecanum robot platform in our lab. Experimental results show that the model is 90% accurate. The results of energy consumption modeling can help robots to save energy by helping them to perform rational path planning and task planning.

## I. INTRODUCTION

The Mecanum wheel is the most common omnidirectional wheel structure used in practical applications. Compared with other omnidirectional wheel structures, it has the advantage of high complexity capability [1]. Mecanum wheel robots have the advantages of stability, high load capacity, simple structure, and flexible move-ment. Therefore, they are widely used in industrial production [2]. While Mecanum wheel robots have the ad-vantage of movement flexibility, they also consume more energy than ordinary wheeled robots [3].

In recent years, Mecanum wheel robots are widely used in warehousing and logistics. With the increasing de-gree of automation of mobile robots, the energy model of mobile robots is becoming more and more complex. How-ever, the energy modeling of Mecanum wheel mobile robots in complex environments have not been given enough attention. In this paper, we mainly study the power consumption modeling of the Mecanum wheel robot in the upslope and downslope environment. A separate analysis of the effects of various factors on power consumption is presented, and a power consumption model (FCEPM) in a complex

The structure of the four-wheeled Mecanum Wheel Robot (FMWR) (see fig. 1) gives it the ability to move omni-directionally and with high loads. The particular structure of the Mecanum wheel provides the robot with better flexibility than ordinary wheeled robots. The omnidirectional movement performance enables mobile robots to car-ry out better transportation tasks in crowded, narrow, or highly dynamic environments. However, the special structure and movement of the Mecanum wheel also increase the energy consumption of the robot, so compared to ordinary wheeled robots, the Mecanum wheel robot has more energy consumption. [5]. Almost all wheeled robots need the energy from the batteries they carry [6]. The energy situation of the battery dramatically limits the task completion of the mobile robot. Energy-efficient strategies for mobile robots can help expand the range of uses, per-form more missions, and accomplish more complex operations [7-9]. Energy modeling has significance for battery energy management and range estimation of mobile robots [10-12]. Accuracy energy models can help robots with task planning, energy prediction [13-17], sensor setup [18-19], and energy optimal path planning [20-22], and also have very significant applications in autonomous planning, autonomous exploration [23], and even robot design.

At present, the main research on the energy consumption of Mecanum wheel robots focuses on the energy consumption of the smooth-running phase of the Mecanum wheel robot during the motion on the plane, the energy modeling in this section has been very well developed[24,25]. However, there is a lack of analysis of the difference in energy consumption of the Mecanum wheel robot between different terrains, and there is no modeling analysis of the factors affecting the energy consumption of the robot, and the fluctuations in energy consumption are only briefly described[25]. This results in incomplete modeling scenes, and the model cannot adapt to the various environments in which the robot operates.Moreover, the accuracy of the modeling needs to be improved.

The main contribution of this paper is to analyze and model the energy consumption of a Mecanum wheel robot in complex environments, the analysis of the factors that have a significant impact on the robot's energy consumption, which are represented as parameters in the modeling. Finally, the difference between the power consumption of Mecanum wheel robots and ordinary wheeled robots due to their special structure is emphasized. According to the previous study, we divided the energy consumption of the Mecanum wheel robot into three major systems: control system, motion system, and sensing system modeled the energy consumption of each of these three systems, and analyzed the interactions and connections between the energy consumption of the three systems. In this paper, we focus on the energy consumption fluctuations of robots, which are often neglected in previous energy consumption studies. We investigate the energy consumption of the Mecanum wheeled robot during uphill and downhill slopes., analyze the differences between the Mecanum wheel robot and ordinary wheeled robots during its motion due to its special structure, and study the fluctuations of energy consumption during the transition phase of its motion on different surfaces and the reasons for them.

The main objective of the study is to increase the applicability of the Mecanum wheel robot modeling as well as to improve the accuracy of the Mecanum wheel modeling to help the Mecanum wheel robot to predict its own energy consumption accurately. By modeling the energy consumption of the Mecanum wheel robot in complex terrain and adding the analysis of the factors affecting the energy consumption of the Mecanum wheel robot, the use scenarios for modeling the energy consumption of the Mecanum wheel can be further increased, and the analysis of the factors affecting the energy consumption can help to effectively improve the accuracy of the modeling. The modeling results can help the robot find the path with the lowest energy consumption during path planning and reduce the robot's energy consumption as well as help the robot predict the energy required for the task and perform reasonable task planning.

## RELATED WORKS

At present, there are many known studied on power consumption models of omnidirectional mobile robots, including omnidirectional wheels and Mecanum wheels robots, terrain and trajectory planning and so on. Sedat Dogru and Lino Marques [1] modeled the energy consumption of the skid steer robot throughout its motion and analyzed each of the factors affecting the robot's energy consumption individually.B. K. Kim et al. [7,26-28] focused on the optimal energy consumption trajectory planning strategy for a three-wheeled omnidirectional wheel robot, proposed that the energy consumption of the robot can be reduced by reducing the speed change of the robot, and planned the lowest energy consumption path for the three-wheeled omnidirectional wheel robot by kinematic and dynamic modeling and combined with the Pontryagin minimax principle, which was verified to save about 30% energy consumption than the ordinary path planning algorithm. Xie. [29] investigated the trajectory planning method for the minimum energy consumption of the Mecanum wheel robot. The method used is to apply the established energy consumption model of the Mecanum wheel to the trajectory planner based on the extended dynamic window method. According to the verification, it can meet the purpose of reducing energy consumption while achieving the online obstacle avoidance function. However, in the study of modeling energy consumption of the Mecanum wheel robot, only the energy consumption of the motion system was modeled, neglecting the energy consumption of the control system and sensing system.

Researches are also evolving rapidly in numerous specific scenarios. Amir Sadrpour et al. [30] examined the problem of mission prediction and battery energy prediction for unmanned ground-operated vehicles using batter-ies as energy sources that undertake some specific tasks. Jesús Morales et al. [31] modeled the energy consumption of a skid steer robot on the hard ground through the two perspectives of power generated by rigid terrain and pow-er provided by the motor. Broderick et al. [19] found that for small and lightweight robots, the energy consumption of the robot's non-motor systems (e.g., sensing, communication, and computing) accounted for a large percentage of the robot's energy consumption, therefore, appropriates scheduling strategies, as well as advanced energy conser-vation techniques and energy-efficient materials, can have a major effect on reducing the energy consumption of robots [32]. Structures such as wheeled robots with redundant brakes [33], limb robots [34] and snake robots [35] that can replace the energy consumption of the motion system are also being investigated. Motor resistances have been identified as the main source of power dissipation in the traction system of wheeled robots, which can be minimized with an appropriate velocity profile [26]. Brateman et al. [19] pointed out that energy consumption can be effectively reduced by coordinating the relationship between the robot motor speed and the processing frequency of the processor and the scanning frequency of the sensors. Power demanded by one motor is almost independent of the speed commanded to the other, whereas in skid-steer, the power required by one motor heavily depends on the speed of the other [36].

## ENERGY MODELING

In order to improve the accuracy of robot modeling in complex environments, we divide the power consump-tion of robots into three major systems, namely motion system, control system and sensing system. Modeling and analyzing the energy consumption of these three systems separately. Finally, the connection between the power consumption of the three systems is sought.

### 2.1 Motion System

The first is that the power consumption of the motion system conforms to the following formula. [1]

$$E_{motion} = \int P_{motor} dt = \int I_a E_m dt$$

$E_{motion}$ is the energy consumed by the robot's motion system, $P_{motor}$ is the total power of the robotic motion system, $I_a$ is the total current flowing through the robot's motion system. $E_m$ is the electromotive force on the robot motor.

The power consumption of the robot's motion system is the power consumption of the robot's DC motor, but the measurement of the current of the robot's motor is a complicated problem to determine , because the currents around the motor are not the same, and the current of the motor of the robot is not the same at different times, so it is very inaccurate to measure the power consumption of the robot motor in this way. Therefore, we decompose the power consumption of the motor and divide the power consumption of the motor into consumption at the place.

$$P_{motor} = P_{colo} + P_{lolo} + P_{output}$$

$P_{motor}$ is the power of the robot's motor, $P_{colo}$ is the power of the robot's copper consumption, $P_{lolo}$ is the power consumption of the robot's iron loss, $P_{output}$ is the mechanical output of the robot.

$$P_{colo} = I_a^2 R_a$$

$P_{colo}$ is the power of the motor to lose energy when the robot is in motion, $I_a$ is the current flowing through the motor, $R_a$ Is the internal resistance of the robot motor.

For a DC motor, the torque $T$ and the rotation speed of the shaft $\omega_{shaft}$ are proportional to the armature current $I_a$ and the electromotive force $E$. At the same time, the torque and speed output of the motor are proportional to the force $F_{wheels}$ and the angular velocity $w$ of the wheel. $\tau$ is the conversion factor of force and energy consumption, which is related to the radius of the wheel and the connection of the wheel to the motor. $k(T)$ is the speed energy factor of the robot $\beta(T)$ is the temperature energy consumption coefficient of the robot, and their values as well as the correspondence with speed and temperature are given in Equation 30 and Equation 31 below.

$$P_{colo} = \tau \cdot F_{wheel}^2 + k(T) + \beta(T) * t$$

$$P_{lolo} = P_{edlo} + P_{hylo} + P_{other}$$

The iron loss of the robot was divided into three parts, where $P_{edlo}$ is eddy current loss, $P_{hylo}$ is hysteresis loss, $P_{other}$ is the iron loss after removing eddy current loss and hysteresis loss of other parts of the loss. Because the eddy current loss and hysteresis loss of the robot cannot be measured, the iron loss of the robot is directly related to the speed, acceleration and running time of the robot

$$P_{lolo} = \lambda_1 \sum_{i=1}^{4}(v_i * a) + \lambda_2 (t * a_i * v^2) + \lambda_3 (t^2 * a * v^2)$$

$\lambda_1$, $\lambda_2$ and $\lambda_3$ are the energy consumption coefficients of the robot motors, $v_i$ is the real-time speed of each wheel of the robot, $v$ is the actual motion speed of the robot, $a_i$ is the real-time acceleration corresponding to each wheel of the robot. According to the characteristics of the Mecanum wheel robot, the mechanical output of the robot can be subdivided into the mechanical energy of the output and the power consumption due to the friction losses caused by the characteristic structure of the Mecanum wheels. which is:

$$P_{output} = P_{mechanical} + P_{fica}$$

Through the principles of motor science, we can learn that the motor's mechanical output is transformed into the torque $T$ and rotational speed $\omega_{shaft}$ of the motor shaft, and they are proportional to the input current $I_a$ and input voltage $E_b$ of the motor. For the robot, the output force $F_{wheels}$ of the wheels is proportional to the torque $T$ of the motor, while the rotational speed of the wheel is equal to the product of the torque $\omega_{shaft}$ of the motor and the diameter of the wheel [2].

$$P_{mechanical} = k_1 F_{wheels}^2 + k_2 \omega F_{wheels}$$

Where $k_1$ and $k_2$ are proportionality constants, which represent information about the motor's torque, armature resistance and other parameters [2].

$$P_{fica} = \mu_1 \sum_{i=1}^{4}(N_i \cdot v_i)\cos\theta + \mu_2 N \cdot v \cos\theta \cos\gamma$$

where $\mu_1$ and $\mu_2$ are the coefficients of friction of the robot wheels, their values and variation laws are given in Equation 25 and Equation 26. $N_i$ is the positive pressure acting on each drive wheel, $\theta$ is the angle between the direction of robot motion and the positive direction of the robot, $\gamma$ is the angle between the inclined plane and the horizontal plane of the robot movement

THE INFLUENCE OF VARIOUS FACTORS ON THE POWER CONSUMPTION OF THE ROBOT

*Research on the power consumption of robots uphill and downhill*

The upslope of the robot mainly affects the speed and acceleration of the robot, which has a huge impact on the robot's control system, motion system and sensing system. The influence on the control system and the sensing system is mainly reflected in the change of the corresponding part of the robot's power consumption due to the change of the speed, and the speed change of the robot can be obtained by substituting it into the formula.

The following is mainly to analyze the power consumption of the motion system of the robot motion system during the ascending process. After analyzing the robot's driving force rise and speed change caused by the angle problem during the uphill process, the impact on the iron loss is reflected in the speed. We do not conduct detailed analysis, and analyze the changes in iron loss and mechanical output.

$$F_{up} = F_{gravity} + F_{friction}$$

$$F_{gravity} = mg \sin\theta \sin\gamma$$

$$F_{friction} = \mu mg \cos\gamma$$

where $F_{friction}$ is the magnitude of the total frictional force on the robot, $\mu$ is the friction factor of the robot, its value is related to both the magnitude of the static friction and the magnitude of the sliding friction of the robot, the specific values and variation laws are given in Equation 24. $m$ is the quality of the robot, $F_{gravity}$ is the component of the robot's gravity along the slope direction, $F_{up}$ is the force that the robot needs to provide during the uphill climb up the slope. Because the robot is subjected to frictional forces down the slope and the downward component of gravity along the slope, the robot must provide a corresponding upward force along the slope to ensure smooth robot operation.

Because it exists during the operation of the robot:

$$F_{up} = F_{wheels}$$

**The effect of temperature on the power consumption of the robot**

Affected by the ambient temperature in the three major systems of the robot are the copper consumption in the robot's motion system and the mechanical output of the robot

As the robot moves, part of the energy of the motor will be converted into heat to be lost. As the temperature of the motor increases, the torque and speed constant of the robot will decrease [37]. To compensate for the decreased torque it is necessary to increase the current input to the motor. Since both $I_a$ and $R_a$ in formula (3) increase, the temperature of the robot will continue to rise. Therefore, if the heat cannot be dissipated reasonably, the robot will lose more and more energy. At the same time, the increased current flowing through the motor drive board will cause the energy consumed by the motor drive board to increase accordingly, and the resulting speed change will cause the control system to lose more and more energy.

The change of resistance with temperature conforms to the formula:

$$R(T) = R_0 \left(1 + \alpha (T - T_0)\right)$$

Among them, $R(T)$ represents the resistance value of the robot when the temperature is $T$, $\alpha$ is the temperature coefficient, $T$ represents the temperature at the current time, $R_0$ and $T_0$ represent the resistance value at a certain time and the temperature at this time.

**Table 1**. Friction constants as measured through different wheels

|  | Front | Back |
|---|---|---|
| M1(Left motor) |  |  |
| $k_{10}$ | 0.0019 | 0.0021 |
| $k_{20}$ | 1.5223 | 1.5325 |
| M2(Left motor) |  |  |
| $k_{10}$ | 0.0022 | 0.0026 |
| $k_{20}$ | 1.5445 | 1.5326 |

SIMULATION AND EXPERIMENTAL VERIFICATION

The experimental setting is mainly to study the power consumption of the robot in various situations during uphill and downhill. Before the experiment we set up a test bench that can freely adjust the uphill and downhill angles. In order to ensure that the power of the robot will be stable on the test bench, during the test we set the slope of the test bench to 4m. The robot we used in the experiment is a four-wheeled Mecanum wheel robot called minibalance in our laboratory. The main parameters of the robot are the mass of the robot: $m = 12kg$, the overall dimensions of the robot are length: $L_1 = 0.86m$, width: $L_2 = 0.52m$, the motor of the robot is the planetary gear motor model MD36N, the design of the robot is independent suspension, which has good shock absorption function and can ensure that the robot can still be used normally in the process of shifting the center of gravity, the maximum load weight of the robot is 30kg. The robot is designed with 24v power supply and the control board of the bottom control system is stmf407.

During the experiment, we increased the height of the test bed continuously, and used German-made inclinometer (the accuracy of the inclinometer can reach $0.05°$) for accurate angle measurement, then control the FMWR to start from the flat bottom first, and then go uphill when the speed is stable. This process requires PID to intervene to ensure that the speed of the robot remains constant during the uphill. When the robot stabilizes in the uphill process, it immediately enters the horizontal movement stage to buffer, and then enters the downhill stage. During the downhill process, the robot control system needs to control the speed of the robot to travel at a preset speed to prevent out of control situations. In order to study the influence of various factors in the environment on the power consumption of the robot, we use different speeds and temperatures and the center of gravity to perform the same experimental process during the experiment. We verify the correctness of our model by comparing the differences in power consumption.

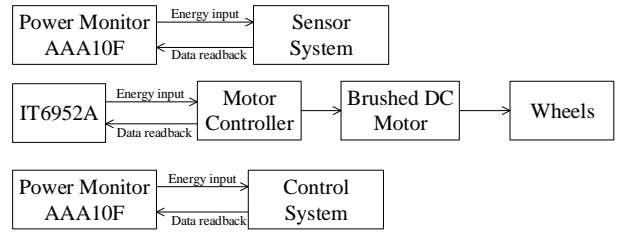

**Figure.1.** Schematic diagram of FMWR downhill stress analysis

The measurement method adopted by the robot is shown in Figure 5. For the control system and the sensor system, the power is less than 10W and the sampling value of the power consumption is relatively high. We use Power Monitor AAA10F from Monsoon Company in the United States to measure power consumption. For motion system because of its high power, we use IT6952A from ITECH Company in the United States to measure. The actual measurements shown in the article are averaged after several is experimental measurements, and we operate in accordance with statistical principles. This ensures that the results of the experiments are not subject to chance and guarantees the accuracy of the experimental validation as well as the reproducibility of the experiments.

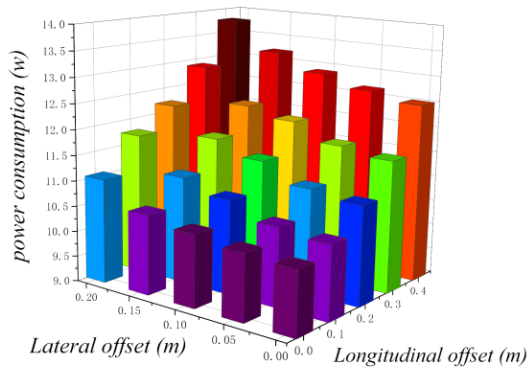

**Figure.2.** Influence of the shift of the center of gravity on the power consumption of the robot

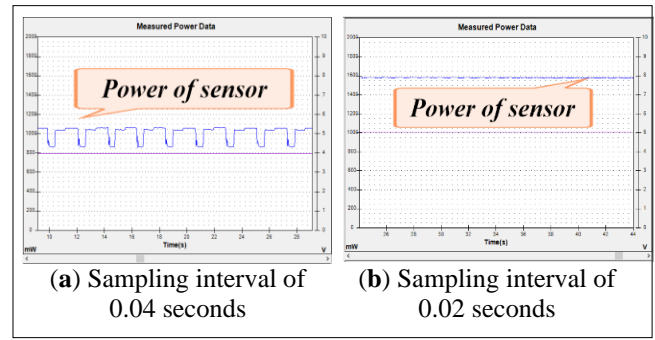

(**a**) Sampling interval of 0.04 seconds    (**b**) Sampling interval of 0.02 seconds

**Figure.3.** Power of the sensor system.

DISCUSSION AND CONCLUSION

Figure 2 shows the effect of the robot's center of gravity offset on the position of the robot's center of gravity on the power consumption of the robot. The data in Figure 9 are the actual measurements of the effect of the change in the robot's center of gravity on energy consumption. In order to show the results more concisely, we only show the actual measured values here, and the error between the simulated and actual values does not exceed 0.5w after the actual comparison.

For the shift of the center of gravity, the method shown in Figure 1 (a) is adopted, the position of the center of gravity of the robot is shifted by placing heavy objects at different positions of the robot. Figure 2 shows the position of the center of gravity of the robot gradually shifted outward from the position of the center of the robot, through Figure 2, it can be seen very clearly that the power consumption of the robot increases significantly as the center of gravity of the robot shifts outward. As the center of gravity of the robot shifts outward, the load on one or more wheels of the robot will increase significantly, which will lead to the increase of power consumption of the corresponding motor of the robot. Moreover, the increase of power consumption is not a linear relationship with the movement of the robot's center of gravity, but a feature of exponential increase, that is. The farther the robot's center of gravity is from the center, the more energy consumption is increased for the same outward movement of the robot's center of gravity. And this kind of shift of the center of gravity is very likely to cause the overload situation of the robot motor.

Figure 3 shows the robot sensing system's power consumption at different sampling frequencies. Figure 3(a) shows the energy consumption of the sensing system when the robot sampling frequency is 0.04 s. Figure 3(b) shows the energy consumption of the sensing system when the robot sampling frequency is 0.02 s. According to the graph, it can be concluded that the higher the sampling frequency of the sensor, the higher the energy consumption of the sensing system. According to the setting the faster the speed of the robot, the higher the sampling frequency will be, and by generalizing it, we can get that the faster the speed of the robot, the higher the energy consumption of the sensing system. We have demonstrated the energy consumption characteristics of the sensor system in the previous research article [24,25], so we will not elaborate on them here.

By comparing the modeling with actual values, it is found that the accuracy of the robot modeling is about 95%, and the modeling has reached the expected modeling accuracy. The difference between the modeling and the actual value is mainly because the consumption of the robot's other parts of the robot's motion system during the modeling process has not been thoroughly searched.

This paper proposes a power consumption model of a robot in a complex environment. The influences of the angle of the upslope and downslope of the robot during the movement, the temperature of the robot and the change of the center of gravity of the robot on the power consumption of the robot are fully studied. Compared to traditional modeling, we pay more attention to the fluctuation of the power consumption of the robot during the transition of the different motion states of the robot. By incorporating the fluctuations of the robot into the modeling, we improved the accuracy of the robot modeling to about 95%, which greatly improved the accuracy of the robot modeling. At the same time, we analyzed the state of Mecanum wheels during actual operation and found that Mecanum wheel robots and ordinary wheeled robots are different in the process of uphill and downhill. The angle limit of the Mecanum wheel robot on uphill and downhill is much smaller than that of the ordinary wheeled robot. This is due to the particular structure of Mecanum wheels, that is, the Mecanum wheels robot sacrifices the robot's ability to adapt to complex environments to increase the flexibility of the robot.